\documentclass[letterpaper, 10 pt, conference]{ieeeconf}  %

\IEEEoverridecommandlockouts                              %

\usepackage[hidelinks]{hyperref}
\usepackage{wrapfig}
\usepackage{times}
\usepackage{amsmath} 
\usepackage{multicol}
\usepackage{graphicx}
\usepackage{amssymb}
\usepackage{booktabs}
\usepackage{multirow}
\usepackage{subcaption}
\usepackage[export]{adjustbox}
\usepackage{wrapfig,lipsum,booktabs}
\let\labelindent\relax
\usepackage{enumitem}

\usepackage{color}
\definecolor{turquoise}{cmyk}{0.65,0,0.1,0.3}
\definecolor{purple}{rgb}{0.65,0,0.65}
\definecolor{dark_green}{rgb}{0, 0.5, 0}
\definecolor{orange}{rgb}{0.8, 0.6, 0.2}
\definecolor{red}{rgb}{0.8, 0.2, 0.2}
\definecolor{darkred}{rgb}{0.6, 0.1, 0.05}
\definecolor{blueish}{rgb}{0.0, 0.3, .6}
\definecolor{light_gray}{rgb}{0.7, 0.7, .7}
\definecolor{pink}{rgb}{1, 0, 1}
\definecolor{greyblue}{rgb}{0.25, 0.25, 1}

\definecolor{PZH_color}{RGB}{0, 102, 204}
\definecolor{TSH_color}{RGB}{218, 126, 63}

\usepackage{xcolor}

\newcommand{\zz}[1]{\textcolor{teal}{}}

\newcommand{\citet}[1]{\cite{#1}}
\newcommand{\citep}[1]{\cite{#1}}

\usepackage[capitalize]{cleveref}
\crefname{section}{Sec.}{Secs.}
\Crefname{section}{Section}{Sections}
\Crefname{table}{Table}{Tables}
\crefname{table}{Table}{Table}
\crefname{figure}{Figure}{Figures}
\crefname{figure}{Fig.}{Figs.}
\Crefname{algocf}{Algorithm}{Algorithms}

\usepackage[noend]{algorithm2e}
\usepackage{algcompatible}%

\newcommand{\hrulealg}[0]{\vspace{1mm} \hrule \vspace{1mm}}

\newcommand{\expect}{\mathop{\mathbb E}}%

\title{
Data-Efficient Learning from Human Interventions for Mobile Robots
}

\author{Zhenghao Peng$^{1}$, Zhizheng Liu$^{1}$, Bolei Zhou$^{1}$%
\thanks{$^{1}$ Department of Computer Science,
        University of California, Los Angeles,
        {\tt\small \{pzh,zhizheng,bolei\}@cs.ucla.edu}}%
}

\begin{document}

\maketitle

\thispagestyle{empty}
\pagestyle{empty}

\begin{abstract}
Mobile robots are essential in applications such as autonomous delivery and hospitality services. Applying learning-based methods to address mobile robot tasks has gained popularity due to its robustness and generalizability. Traditional methods such as Imitation Learning (IL) and Reinforcement Learning (RL) offer adaptability but require large datasets, carefully crafted reward functions, and face sim-to-real gaps, making them challenging for efficient and safe real-world deployment.
We propose an online human-in-the-loop learning method \textit{PVP4Real} that combines IL and RL to address these issues. PVP4Real enables efficient real-time policy learning from online human intervention and demonstration, without reward or any pretraining, significantly improving data efficiency and training safety. We validate our method by training two different robots---a legged quadruped, and a wheeled delivery robot---in two mobile robot tasks, one of which even uses raw RGBD image as observation. The training finishes {within 15 minutes}. Our experiments show the promising future of human-in-the-loop learning in addressing the data efficiency issue in real-world robotic tasks.
More information is available at \url{https://metadriverse.github.io/pvp4real/}.
\end{abstract}

\section{Introduction}
\label{sec:intro}

Mobile robots have become increasingly important in a variety of applications, ranging from autonomous delivery~\cite{engesser2023autonomous} in urban environments to service robots in healthcare~\cite{morgan2022robots} and hospitality~\cite{tuomi2021applications}.
\begin{figure}[!t]
\centering
\includegraphics[width=\linewidth]{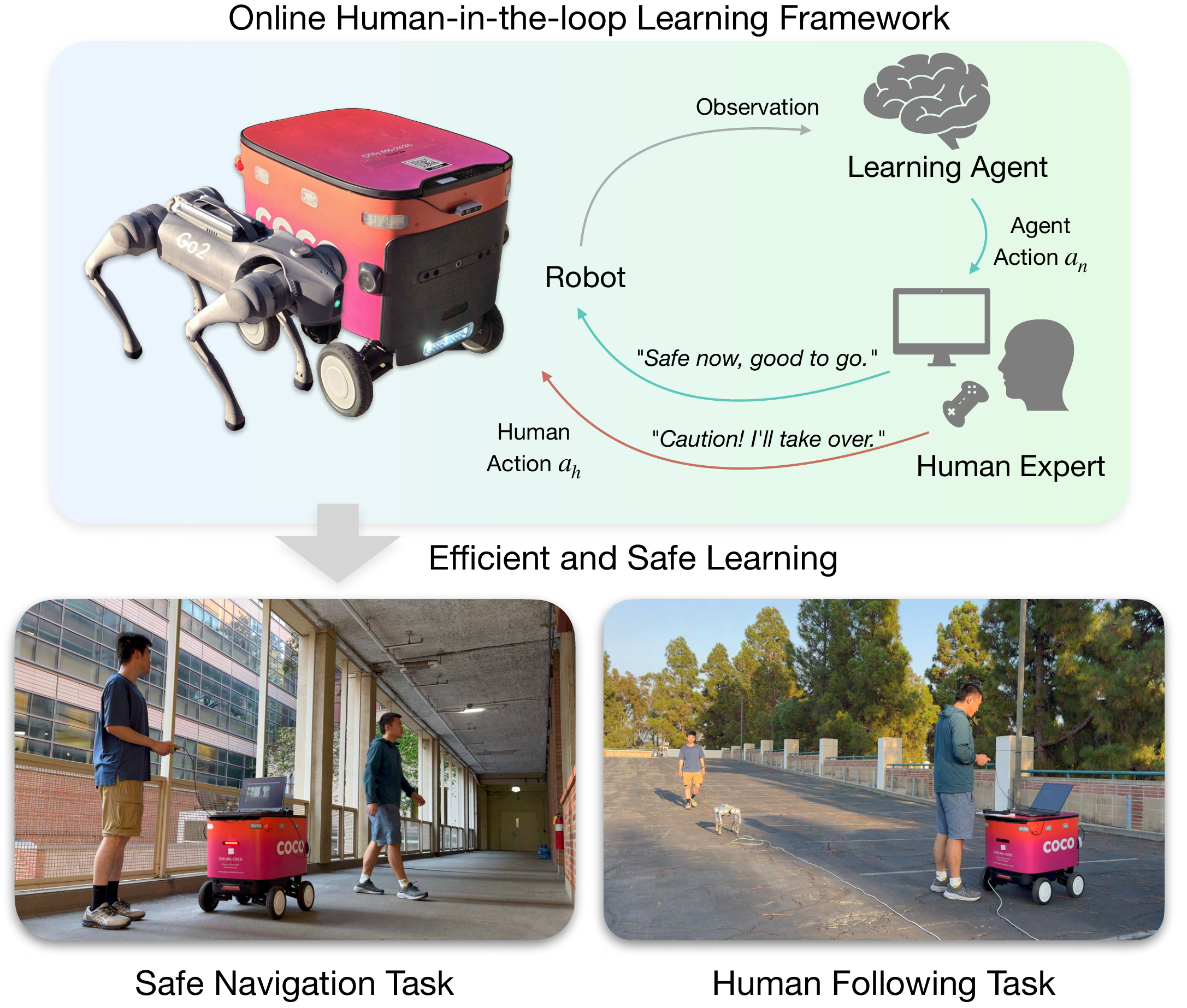}
\caption{
We train mobile robots in dynamic environments with a human-in-the-loop learning method.
We can solve challenging tasks with raw camera input in as little as 15 minutes, training from scratch, without reward and only learning from online human-in-the-loop intervention and demonstration.
}
\vspace{-1em}
\label{fig:teaser}
\end{figure}
Recently, learning-based methods have gained popularity for their robustness and adaptability by allowing robots to learn from experience rather than relying on pre-defined rules and models of the environment. Two common learning-based approaches are Imitation Learning (IL) and Reinforcement Learning (RL). IL trains robots by mimicking human demonstrations~\cite{pan2020zero,hoeller2021learning,seo2023learning,brohan2023rt,sridhar2024nomad}, allowing them to learn complex behaviors directly from experts. However, large amounts of demonstration data are required, which is expensive and time-consuming to collect. IL methods also suffer from compounding errors due to the distribution shifts between the training data and the real-world environments, which is more prominent when the data is insufficient.
On the other hand, RL involves training robots through trial and error by interacting with the environment and optimizing a reward function. RL methods often depend on carefully engineered reward functions, which can be challenging to design and tune for complex, real-world tasks. 
Also, the learned agent may obtain biased behaviors or figure out the loophole to finish the task~\citep{leike2018scalable,russell2019human, pan2022effects}.
Additionally, there is the sim-to-real gap problem, where policies trained in simulation do not transfer well to the real world due to the noisy sensory data and differences in the environments. Furthermore, training RL models directly in the real world is risky because it requires extensive exploration that could lead to unsafe actions, such as collisions with surroundings.
Given these challenges, \textbf{there is a strong need for more data-efficient and safer learning methods for robotic training in the real world}.

Human-in-the-loop approaches, which involve active human intervention and demonstration during training, provide a promising alternative.
Instead of generating an offline dataset and training novice policy against it~\cite{ho2016generative,fu2018learning}, we can incorporate a human subject into the training loop to provide online data. This can mitigate the distributional shift since the data generated with human-in-the-loop has closer state distribution to that of the novice policy, compared to the data entirely collected from human demonstration~\cite{ross2010efficient}.
These methods can significantly reduce the required data and ensure safety during training by allowing humans to provide corrective demonstrations in real-time, guiding the robot toward safe and preferable behaviors.
For safety-critical scenarios in which mobile robots are commonly used, safety is undoubtedly the priority.
Therefore, instead of collecting human feedback after robot's interactions with the environment and then learning from these preference feedback~\cite{wirth2017survey,christiano2017deep,reddy2018shared, warnell2018deep, palan2019learning,guan2021widening,ouyang2022training}, we opt for \textbf{online human-in-the-loop learning}, where human subjects actively intervene and provide demonstrations during training~\cite{kelly2019hg,spencer2020learning,mandlekar2020human,li2021efficient,peng2023learning}.
With online correction and demonstration from human subjects, AI alignment and training-time safety of the system can be substantially enhanced.
However, most previous works focus on the simulation environment~\cite{li2021efficient,peng2023learning,huang2024human}. Whether the human-in-the-loop method is applicable to the real world and can overcome challenges such as noisy sensory reading, teleoperation delay, and the complexity of visual appearance and robot kinematics has not yet been examined.

In this work, we develop a data-efficient policy learning method \textit{PVP4Real} to harness online human interventions and demonstrations. It can be generalized across different robots, tasks, and input representations in the real world. As shown in \cref{fig:teaser}, we showcase that we are able to train two different mobile robots in two challenging tasks with high data efficiency.
In the safe navigation task, the robot needs to move forward while avoiding collision. The human subject might jump in suddenly, so the robot needs to learn to emergency stop.
In the human following task, the robot receives the bounding box input of a tracked person and must follow the human while keeping away at a certain distance and stop if the human stops. The robot must be responsive and agile, otherwise the tracked person might move out of the view. 
Those two tasks are well fitted to the online human-in-the-loop learning, as both tasks are human conductible, meaning humans can give control signals with the user interface. Meanwhile both tasks suffer from the huge sim-to-real gap as it is hard to simulate noisy camera images and complex human behaviors.
\textbf{PVP4Real can train robots to solve these tasks {within only 15 minutes in wall time.}}
Behind the successful experiments is an online human-in-the-loop learning algorithm, which combines the RL and IL and learns from human interventional data and online demonstration. 
Even though we adopt a reward-free setting, the TD learning of RL helps the model generalize learned human preference across different state spaces~\cite {lu2023imitation,chu2025sft}. The IL component constrains the policy to be closed to the data distribution seen before and regularizes the policy~\cite{fujimoto2021minimalist}.
We summarize our main contributions as follows:
1)
We propose PVP4Real, a data-efficient human-in-the-loop learning method. Our method is \textit{reward-free} and can be generalized across various task settings, sensory data, and robotic embodiments.
2)
We build real-world experiment systems with two robots and deploy the proposed method directly on the systems. 
The experiments show that the proposed method enables superior performance and high learning efficiency in two important mobile robot tasks: safe navigation and human following 
in just 15 minutes, training from scratch without reward and prior knowledge.

\section{Related Work}
\label{section:related-works}

\subsection{Human-in-the-loop policy learning}

Online human-in-the-loop learning allows human subjects to proactively involve the agent-environment interactions to ensure safety.
Human subjects can intervene in the episode if a near-accidental situation happens, and such intervention policy can be learned~\citep{zhang2017query,abel2017agent,saunders2018trial,pakdamanian2021deeptake,xu2022look,wang2021appli}. 
Recent studies explore learning from both intervention and demonstration in the human-agent shared autonomy setting~\citep{macglashan2017interactive,menda2019ensembledagger,kelly2019hg,spencer2020learning,li2021efficient,jonnavittula2021learning,xu2022look,luo2023rlif}.
However, previous methods do not fully utilize the power of human involvement.
Interactive imitation learning method (HG-DAgger)~\citep{kelly2019hg} does not leverage data collected by agents, while Intervention Weighted Regression (IWR)~\citep{mandlekar2020human} does not suppress undesired actions likely intervened by humans.
Meanwhile, Expert Intervention Learning (EIL)~\citep{spencer2020learning} and IWR~\citep{mandlekar2020human} focus on imitating actions step-wise without considering the temporal correlation between steps.
Our earlier work, Proxy Value Propagation (PVP)~\cite{peng2023learning}, learns a proxy value function from the human intervention and demonstration without explicitly regularizing the policy to be closed to human behavior. 
These drawbacks harm learning efficiency and thus incur more human involvement. 
Most of the previous works focus on the simulation environment~\cite{mandlekar2020human,peng2023learning,huang2024human},
and \textit{only a few works have directly applied human-in-the-loop online learning to the real-world robots} due to practical challenges such as noisy observation, teleoperation delay, complex visual appearance, and varying robot dynamics.
In EIL~\cite{spencer2020learning}, the learning method has been applied to a simple task: drive a racing robot on a rectangular track without collisions.
In HG-DAgger~\cite{kelly2019hg}, the task is to drive a vehicle on a straight road.
In these works, the observation is a low-dimensional state vector of the location and velocity of the robot.
In contrast, in the safe navigation task we experiment with, a CNN network is trained from scratch with raw RGBD images. Also, our experiments are more complex than the previous works regarding different tasks, embodiments, and input formats.

\subsection{Learning for Mobile Robot}
Mobile robots are designed to navigate and interact autonomously within their environments, necessitating continuous perception, localization, and navigation capabilities. 
With the rapid development of deep learning, many learning-based methods have been introduced in mobile robotics. One line of work adopts a modular design, training separate models for each sub-task---for instance, using object detectors for perception~\cite{ren2016faster}, learning-based SLAM for localization~\cite{mur2015orb}, and reinforcement learning (RL) controllers for navigation~\cite{zhu2017target}. This approach allows for specialized optimization of each component but may suffer from integration challenges and cumulative errors across modules.
Another line of work embraces an end-to-end paradigm that directly maps sensor observations to control actions~\cite{bojarski2016end, pfeiffer2017perception}. End-to-end learning removes the need for handcrafted intermediate representations and potentially captures complex correlations in the data. However, it often requires large amounts of training data and may lack interpretability for error diagnosis.  
Our experiments cover both approaches. In the human following task, we employ a modular setup where an object detector and tracker provide bounding box inputs of the target person, similar to methods used in~\cite{muller2018driving, gervet2023navigating}. Conversely, in the safe navigation task, we adopt an end-to-end learning strategy by training a policy that takes raw RGBD images as input, inspired by works like~\cite{pfeiffer2018reinforced, zhu2017target}. This dual approach demonstrates the versatility of our human-in-the-loop learning method across different robotic embodiments and sensory inputs.

\section{Method}

\begin{figure}[!t]
\centering
\includegraphics[width=\linewidth]{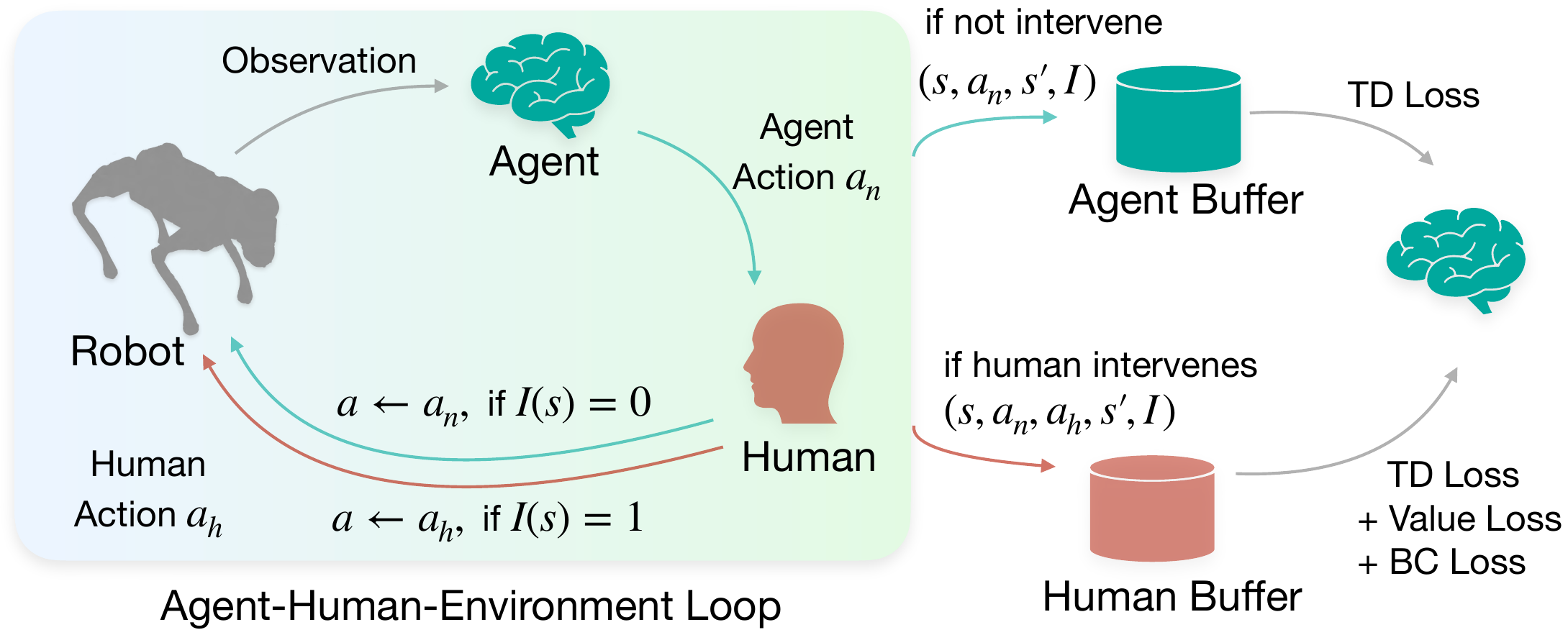}
\caption{
 \textbf{Method Overview.} In human-in-the-loop learning, the human subject supervises the action of the learning agent $a_h$ and decides whether they should intervene and use the human action $a_h$ instead. Depending on the intervention results, we store the transitions in two separate replay buffers. Data from both buffers are sampled to update the policy and value networks in real-time, alongside the ongoing interactions between the agent, human, and environment. No reward and prior knowledge are needed.
}
\vspace{-1em}
\label{fig:framework}
\end{figure}

In this section, we introduce an efficient online human-in-the-loop learning algorithm \textit{PVP4Real} that trains agents from human interventions and demonstrations, without reward or any form of pretraining.

\vspace{-0.2em}
\subsection{Problem Formulation}
\vspace{-0.2em}

Our goal is to train a policy to solve the sequential decision-making problem. This is usually modeled by a Markov decision process~(MDP). 
MDP is defined by the tuple $M=\left\langle \mathcal{S}, \mathcal{A}, \mathcal{P}, r, \gamma, d_{0}\right\rangle$ consisting of a state space $\mathcal{S}$, an action space $\mathcal{A}$, a state transition function $\mathcal{P}:\mathcal{S}\times\mathcal{A}\to\mathcal{S}$, a reward function $r: \mathcal{S}\times\mathcal{A}\to[R_{\min}, R_{\max}]$, a discount factor $\gamma\in(0,1)$, and an initial state distribution $d_0:\mathcal{S}\to[0,1]$.
In reinforcement learning, the goal is to learn a \textit{novice policy} $\pi_n(a | s): \mathcal{S}\times\mathcal{A}\to[0,1] $ that can maximize the expected cumulative return:
$
\pi_n = \arg\max_{\pi_n} \expect_{\tau\sim P_{\pi_n}}[
\sum_{t=0}^{T} \gamma^{t} r(s_t, a_t)],
$
wherein $\tau = (s_0, a_0, ..., s_T, a_T)$ is the trajectory sampled from trajectory distribution $P_{\pi_n}$ induced by $\pi_n$, $d_0$ and $\mathcal P$.
We denote the state-action value and state value of a policy $\pi$ as 
$Q(s,a)=\mathbb{E} \left[\sum_{t=0}^{\infty} \gamma^{t} r\left(s_{t}, a_{t}\right)\right]$
and $V(s)=\mathbb{E}_{a\sim\pi(\cdot|s)}Q(s,a)$, respectively. Note that we will reuse the concept of values in the following, but we are in the \textit{reward-free} setting where the environmental reward is not accessible. The goal of this paper is to learn a human-compatible policy that minimizes human intervention and can best capture human preferences.

Our method approaches policy learning with human-in-the-loop online intervention.
Assuming the human expert has a \textit{human policy} $\pi_h(a_h|s): \mathcal{S}\times\mathcal{A}\to[0,1] $, which outputs human action $a_h \in \mathcal A$, sharing the same action space as novice action.
We can model human intervention by an \textit{intervention policy} $I(s)$ to describe human subjects' interventional behaviors. The intervention is boolean $I\sim I(\cdot|s), I \in \{0, 1\}$.
As shown in \cref{fig:framework}, during training, a human subject accompanies the novice policy and can intervene the agent by taking over the control to demonstrate desired behaviors.
The behavior policy and the behavior action $a$ can be written as:
\begin{equation}
  \pi_b(\cdot|s) 
  =
  (1 - I(s)) \pi_n(\cdot|s)
  +
  I(s) \pi_h(\cdot | s).
\end{equation}

We directly learn decision-making policy with real-world robots, and no reward function is defined. The only data we can collect from agent-human-environment interaction is the transitions $\{(s, a_n, a_h, I, s')\}$ where $s'$ is the next state and $a_h$ might be null if a human does not intervene. Thus, the key challenge to be addressed by this work is how we can effectively utilize these transitions to recover human preferences and obtain novice policy that is well-performing in test time without human involvement.

\subsection{Human-in-the-loop Training}

As suggested by~\cite{russell2019human}, the ultimate source of information about human preferences is human behavior. 
Our first observation is that we can assume human actions to be optimal, and the agent should reproduce those actions when in the same state.
A straightforward idea is to apply Behavior Cloning (BC) loss on human data by minimizing the negative log-likelihood loss:
\begin{equation}
\label{eq:bc-loss}
J^\text{BC}(\phi) = - \expect_{(s, a_n, a_h)} I(s) \log \pi_n^\phi(a_h|s),
\end{equation}
wherein the novice policy $\pi_n$ is parameterized by $\phi$. This resembles the objective of HG-DAgger~\citep{kelly2019hg} and the intervention loss in EIL~\cite{spencer2020learning}.

\RestyleAlgo{ruled}
\begin{algorithm}[!t]
\caption{Online Human-in-the-loop Learning}
\label{algo}
\begin{small}
\SetAlgoLined
\LinesNumbered
\DontPrintSemicolon
\KwIn{
Environment; Minibatch size $N$;
Human subject modeled by $\pi_h$ and intervention policy $I$.
\\
}
\KwResult{
Learned novice policy $\pi_n^\phi$.
\\ \hrulealg
}
Initialize replay buffer $\mathcal D_h, \mathcal D_n$. \\
\For{every $0.2$s}{
Receive observation $s$ from the environment. \\
Query novice action $a_n \gets \pi_n^\phi(\cdot|s) $. \\
Query human intervention $I(s)$. \\
\uIf{$I(s) = 0$}{
    $a_h \gets \emptyset$, $a \gets a_n$. \\
  }
  \Else{
    $a_h \gets \pi_h(\cdot|s)$, $a \gets a_h$. \\
  }
Send $a$ to the robot and get the next $s'$. \\
Store $(s, a_n, a_h, I, s')$ to $\mathcal D_h$ (if intervenes) or $\mathcal D_n$. \\
Sample $N$ transitions from $\mathcal D_n$ and $\mathcal D_h$ seperately and update $\theta, \phi$ by minimizing \cref{eq:our-final-value-loss} and \cref{eq:our-final-policy-loss}. \\
}
\end{small}
\end{algorithm}

The second observation is that human intervention implies that the agent's actions being interrupted is undesirable and needs to be avoided.
Previous works like IWR~\citep{mandlekar2020human} and HG-Dagger do not suppress undesired actions intervened by humans.
EIL~\cite{spencer2020learning} directly modifies the Q values of those actions so that the policy (an argmax policy on Q values) will avoid those actions. As we want to work with mobile robots whose action spaces typically have velocity commands and are continuous, we can incorporate the value iteration in RL literature and first update the Q value, then update the policy to maximize the Q value. Here, we introduce a value network $Q^\theta(s, a) \to \mathbb R$ parameterized by $\theta$. We can let the Q network predict a positive value for the human behaviors and a negative value for those novice actions that are intervened by human subjects, following our previous work PVP~\cite{peng2024learning}:
\begin{equation}
\label{equation:new-proxy-value-1}
    J^\text{Q}(\theta) = 
    \expect_{(s, a_n, a_h)}   [| Q^\theta (s, a_h) - B |^2  +    | Q^\theta (s, a_n) + B |^2 ]I(s),
\end{equation}
wherein $B > 0$ is a constant. $J^\text{Q}(\theta)$ will be minimized.

The third observation is that those agent actions in which humans do not intervene can be considered human-compatible.
EIL proposes to conduct IL on those actions. However, to determine which part of the data IL will use, a carefully tuned hyper-parameter is needed to decide whether the agent's actions will cause human intervention later. Meanwhile, the step-wise IL on those unreliable agent actions might cause compounding errors.
Our insight is that we can utilize this part of the data from the RL perspective.
RL policies can establish causal relationships between observations, actions, and outcomes as they are trained in closed-loop. This makes them less vulnerable to covariate shifts and spurious correlations commonly seen in open loop IL~\cite{lu2023imitation}. Also, those transitions contain information on the forward dynamics of the environment~\citep{levine2020offline,yu2022leverage}.
We propose to conduct TD learning instead of step-wise imitation learning. We can apply TD learning on all data (reward is removed as our method is in a reward-free setting):
\begin{equation}
\label{eq:td-learning}
    J^\text{TD}(\theta) = \expect_{(s, a, s')} | Q^\theta(s, a) -  \gamma \max_{a'} Q^{\hat{\theta}}(s', a') |^2,
\end{equation}
where $Q^{\hat{\theta}}$ is the delay-updated target network. 
We can use $-\expect_{s, a\sim\pi_n^\phi(\cdot|s)} Q^\theta(s, a)$ to update policy as in TD3~\cite{fujimoto2018addressing}. 
By incorporating the RL framework, 
the objective to update the value network becomes:
\begin{equation}
\label{eq:our-final-value-loss}
J^\text{Ours}(\theta) = J^\text{Q}(\theta) + J^\text{TD}(\theta).
\end{equation}
The objective of updating the policy network becomes:
\begin{equation}
\label{eq:our-final-policy-loss}
J^\text{Ours}(\phi) = -\expect_{s, a\sim\pi_n^\phi(\cdot|s)} Q^\theta(s, a) + J^\text{BC}(\phi).
\end{equation}
Note that as suggested by~\cite{fujimoto2021minimalist, lu2023imitation}, adding behavior cloning loss into the RL objective can balance exploiting the data in the dataset and regularizing the policy towards actions seen in the dataset. This benefits human-in-the-loop learning as it makes policy closer to the human policy. %

\noindent \textbf{Implementation Details.}
With human in the loop, the transitions can be categorized into two sets: human-involved transitions and not-involved transitions. We create two replay buffers to store those transitions separately following PVP~\cite{peng2023learning}. Denote $\mathcal D_h$ as the human buffer, which stores transitions $(s, a_n, a_h, I, s')$ when human intervenes $I(s)=1$. Another buffer $\mathcal D_n$, named agent buffer, stores the transitions $(s, a_n, I, s'), I(s)=0$. We will sample the same amount of data from these two buffers to avoid data imbalance in training.
\cref{algo} describes the procedure of our method: we conduct online human-in-the-loop learning and update the policy and value network at each environmental interaction to minimize \cref{eq:our-final-value-loss} and \cref{eq:our-final-policy-loss}.
We implement our method in Stable-Baselines3~\cite{stablebaselines3} and set the minibatch size to N=1024, learning rate 1e-4. The value network and policy network do not share parameters. There is a task-specific feature extractor that outputs a 256-dim feature vector, followed by a two-layer MLP with 256 hidden size. We set $B=1$. The target Q network is delayed and updated by coefficient $\tau=0.05$.
See \cref{section:experiments} for details on real-world robot experiments.

\vspace{-0.2em}
\section{Simulation Experiments}
\label{section:simulation-exp}

\vspace{-0.2em}

\begin{figure}[!t]
\centering
\includegraphics[width=\linewidth]{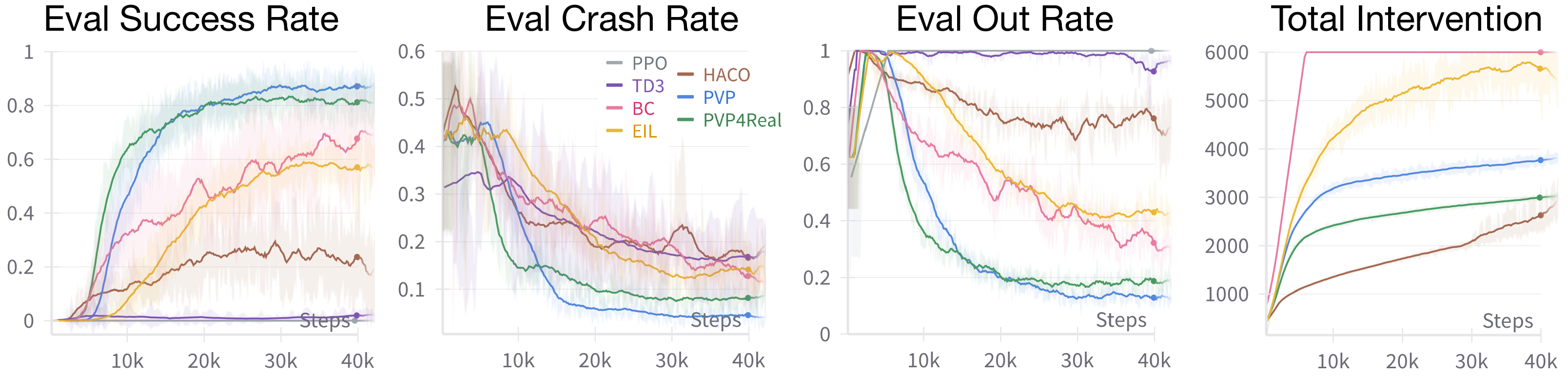}
\vspace{-1.5em}
\caption{
\textbf{Experiment on Simulation Environment.}
Human-in-the-loop methods achieves much better sample efficiency compared to the RL baselines.
}
\vspace{-1.5em}
\label{fig:metadrive-curve}
\end{figure}

We conduct simulation experiments in the MetaDrive Safety Environment~\cite{li2021metadrive} following previous works~\cite{li2021efficient,peng2023learning,huang2024human}.
The task for the agent is to steer the target vehicle with acceleration, brake, and steering to reach the predefined destination. The ratio of episodes where the agent successfully reaches the destination
is called the \textit{success rate}. To increase the task's difficulty, we scatter obstacles randomly in each
driving scene, such as movable traffic vehicles, fixed traffic cones, and warning triangles. 
The proportion of episodes terminated due to collisions with obstacles is called \textit{crash rate}. Similarly, \textit{out rate} is defined by the fraction of episodes where the agent drives out of the road.
The observation is a vector containing current states, navigation information, and surrounding information. We use two-layer MLPs with 256 hidden nodes as the policy and the critic networks.
To simulate the human expert, a PPO~\cite{schulman2017proximal} expert is trained with 20 million environmental interactions. The expert achieves $\sim$0.80 success rate and $\sim$0.10 crash rate. We follow previous works and define a threshold function to automatically issue intervention and provide expert demonstration~\cite{peng2021safe,xueguarded}. We will let the PPO expert take over if the probability of the agent's action under PPO's action distribution is lower than a threshold $I(s) = \mathbf 1_{\pi_\text{PPO}(a|s) < \epsilon}, a\sim \pi_n(\cdot|s)$. We set $\epsilon = 0.05$.
We split the driving scenes into
The training set and test set will have 50 different scenes in each set, and the generalization performance of the learned policy will be evaluated by testing the trained agents in separate test environments.
Per each 200 environment interactions, we roll out the learning agent without the PPO expert in the test environments for 100 episodes to get the evaluation results. 
All experiments are repeated 8 times with different random seeds.
We compare with human-in-the-loop baselines Human-AI Copilot Optimization (HACO)~\citep{li2021efficient}, Expert Intervention Learning (EIL)~\cite{spencer2020learning} and Proxy Value Propagation (PVP)~\cite{peng2024learning}.
For RL baselines, we compared to 
PPO~\citep{schulman2017proximal} and TD3~\citep{fujimoto2018addressing}.
We also compare Behavior Cloning. For BC, instead of pre-collecting a dataset, we let the expert run in the environment and keep expanding the BC dataset.
As shown in \cref{fig:metadrive-curve}, by training on such insufficient 40k environment interactions, RL methods TD3 and PPO hardly learn any useful behavior. BC-trained agents are inferior as the learning policy has a gap to the behavior policy, namely the expert~\cite{ross2011reduction}. On the contrary, human-in-the-loop methods all progress in solving the task. Our method outperforms EIL and HACO in term of task completion while we use less expert intervention. Compared to PVP, our method requires less expert intervention.

\section{Real-world Experiments}
\label{section:experiments}
\vspace{-0.5em}

\subsection{Setup}
\begin{figure}[!t]
\centering
\includegraphics[width=0.8\linewidth]{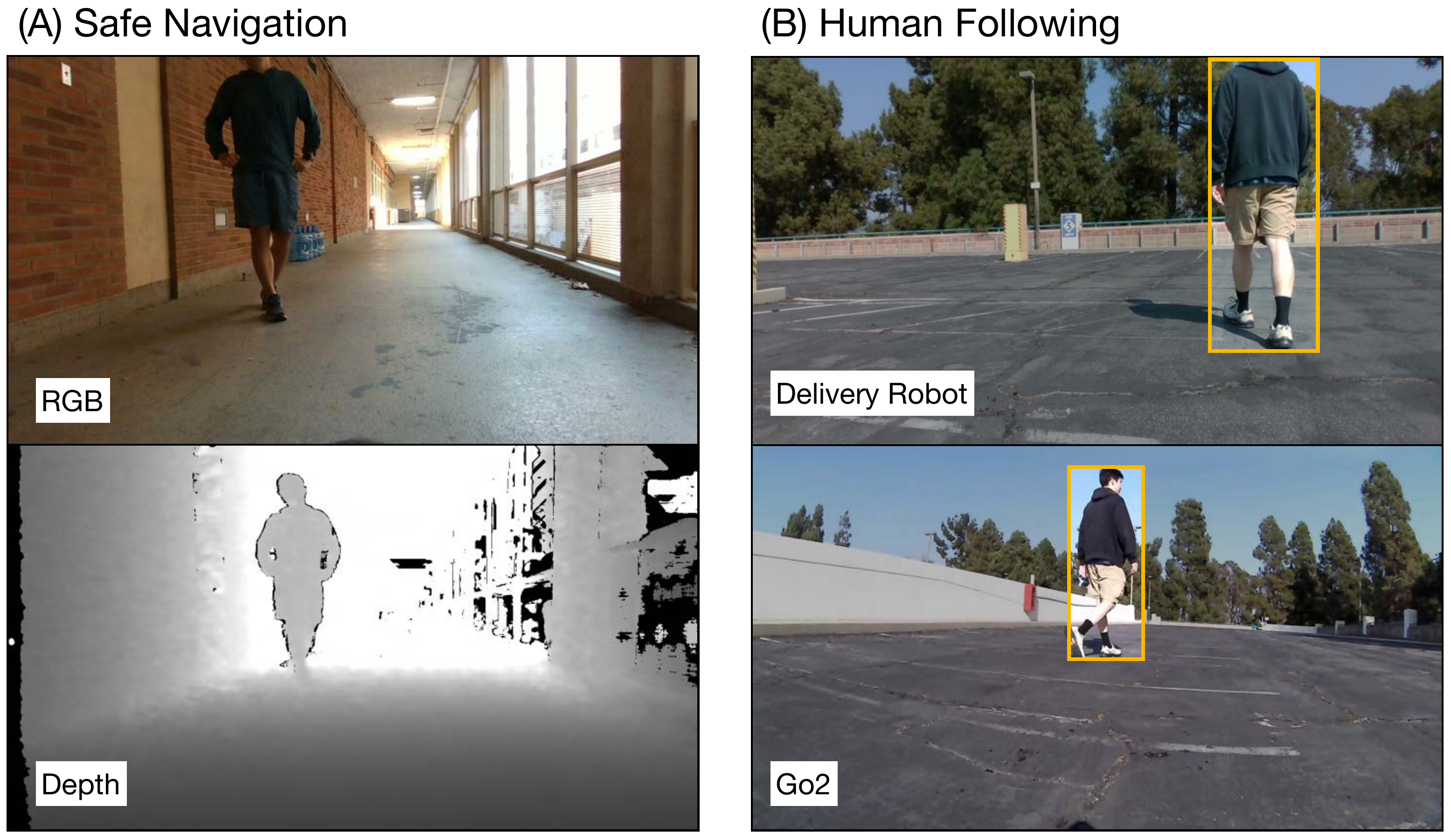}
\caption{
\textbf{Overview of the tasks.}
\textbf{(A)} In Safe Navigation, the agent is given the raw RGB and depth images. The goal is to traverse the corridor environment while avoiding dynamic and moving obstacles, like a walking person, and perform emergency stops if necessary.
\textbf{(B)} In Human Following, the robot is given the 2D bounding box of the tracked human. It needs to follow the human within a certain distance while the human performs different behaviors and keeps the target within the view. We experiment with two embodiments, which have different robot dynamics and camera parameters.
}
\vspace{-1.5em}
\label{fig:observation}
\end{figure}

We conduct experiments on two practical tasks for mobile robots: Safe Navigation and Human Following. An overview of the tasks is shown in \cref{fig:observation}. These tasks feature different settings and input representations to show the generalization ability of our method on real-world applications.

\noindent\textbf{Safe Navigation.}
The robot navigates in a corridor environment by moving forward along the corridor~\cite{seo2022prelude}. The size of the corridor is about 20m in length and 2.5m in width. The environment has static obstacles and moving humans, and the robot needs to avoid collisions while moving forward. The robot is not given any high-level observation, such as its locations, and only takes the RGB and depth images as the input. The robot needs to learn an \textit{end-to-end} visual-motor policy that can safely traverse the scene with only a few human demonstrations without any external reward and prior knowledge.  During training, the environment is randomly initialized with the obstacle positions and moving human trajectories. Some human trajectories would directly intersect with the robot's moving path, so the robot must also perform an emergency stop if necessary. We evaluate the success rate of four different sub-tasks the robot should learn in order to traverse the corridor successfully: 
1) \textit{Moving Straight}: the robot starts from one side of an obstacle-free corridor, and the goal is to reach the other side,  
2) \textit{Static Obstacle Avoidance}: We place static obstacles at an interval of 5m along the path, and the robot succeeds each time it passes an obstacle without collision. 
3) \textit{Dynamic Obstacle Avoidance}: A person walks around the robot at an interval of 5m along the path, and the robot succeeds each time it moves forward without colliding with the person.
4) \textit{Emergency Stop}: A person not in the robot's view suddenly appears in front of the robot and then leaves after 2s. The robot succeeds if it stops immediately when it sees the person and resumes moving after the person leaves. For all sub-tasks, the robot fails if it collides with anything or gets stuck in a position where it should keep moving.
We perform 20 trials for each sub-task and report the average success rate.

\noindent\textbf{Human Following.}
In this task, the robot aims to follow a person within a certain distance ($\sim$4m) in an empty parking lot. It is given the 2D bounding box of the tracked person, provided by an off-the-shelf detector~\cite{liu2023grounding} and tracker~\cite{bhat2019learning}, which run on the RGB images captured by the onboard camera. During training, the target person demonstrates random trajectories, including sudden stops, and we test the trained policy on four specific types of human behavior during evaluation: 
1) \textit{Going Straight}: the person walks in a straight line in the same direction as the robot for 10 meters, then stops.
2) \textit{Following a Curve}: the person moves forward while turning for 10 meters, ending up facing a direction that is perpendicular to where they started. We separately evaluate when the curve is turning left or right. 
3) \textit{Sharp Turn}: the person starts walking in a direction that is perpendicular to the robot for 10 meters. Different from \textit{Following a Curve}, the human walks away immediately after the episode begins. We note that this behavior is never demonstrated during training. 
4) \textit{Sudden Stop}: the person suddenly stops while walking. The robot fails the task if it loses track of the person or is too close ($<$1m) or too far ($>$5m) from the person. 
We perform 20 trials for each type of human behavior and evaluate the average success rate.

\noindent\textbf{Robots.}
We deploy our method in two different embodiments: the Unitree Go2 robot and the four-wheel delivery robot from Coco Robotics, as shown in \cref{fig:teaser}. We use the native RGB camera from the Unitree Go2 to perform human tracking. For the delivery robot, we mount a Realsense D435i depth camera to the front of its chassis for both the task of human following and safe navigation. All cameras are streamed with a resolution of $1280\times720$ in 30 fps.

\noindent\textbf{Baseline.} Our primary baseline is behavioral cloning (BC). We train the policy network in BC with the same amount of human demonstration data ($\sim$2000 steps) as the human-in-the-loop method for 5,000 epochs using the MSE loss.

\noindent\textbf{Implementation Details.}
For both tasks and robots, the action space is 2-dimensional with linear and angular velocities, and we use the robots' low-level controllers to convert the velocity commands to motor torques. 
As for the observation space in the safe navigation task, the images are first resized to 84$\times$84 resolution. We stack the normalized RGB and depth images along the channel dimension and concatenate the images of the past five frames, resulting in the input data in shape $[84, 84, 20]$. 
The feature extractor is a 5-layer convolutional neural network with channels $[16, 32, 64, 128]$.
The observation space of the human following task is 5-dimensional, including the position and dimension of the normalized 2D bounding box of the tracked person and the validity.
We then concatenate the observations of the past five frames into a single vector with 25 dimensions. We use two-layer MLPs with 256 hidden nodes as the policy and value networks.
By stacking the observations from the past 5 steps, the agent is aware of the history information from the past 1s, as the system operates at 5 Hz.
We implement a human intervention interface on a laptop with an Nvidia RTX 4080 mobile GPU to support online training and human control with the joystick. In both tasks, we perform 2,000 steps of environment interactions, which is equivalent to $\sim$7 minutes of human supervision. However, due to the hardware and training overhead, the experiments are completed within 15 minutes.
All experiments with human participants have obtained approval through our institute's IRB.

\subsection{Results}

\begin{table}[!t]
\centering
\caption{\textbf{Success rate of the Safe Navigation task.} }
\vspace{-0.5em}
\label{tab:avoid}
\begin{tabular}{@{}lcccc@{}}
\toprule
 & \shortstack{Moving \\Straight} & \shortstack{Static Obstacle \\Avoidance} & \shortstack{Dynamic Obstacle \\Avoidance} & \shortstack{Emergency \\Stop} \\ \midrule
BC & 0.25 & 0.50 &  0.45 & 0.60 \\
Ours & 1.00 & 0.85 & 0.90 & 1.00 \\ \bottomrule
\end{tabular}%
\end{table}

\begin{table}[!t]
\centering
\caption{\textbf{Success rate of the Human Following task.} 
}
\label{tab:coco-chase}
    \begin{subtable}[t]{\linewidth}
        \centering
        \vspace{-0.5em}
        \caption{Evaluation on the Coco delivery Robot.}
        \vspace{-0.5em}
\begin{tabular}{@{}lcccc@{}}
\toprule
\textit{Coco} & Straight & Left Curve & Right Curve & Sudden Stop \\
 \midrule
BC & 1.00 & 0.60 & 0.60 & 0.45 \\
Ours & 1.00 & 0.80 & 1.00 & 0.85 \\
\bottomrule
\end{tabular}%
\end{subtable}
\begin{subtable}[t]{\linewidth}
\vspace{2mm}
\caption{Evaluation on the Unitree Go2 robot.}
\vspace{-0.5em}
\centering
\label{tab:dog-chase}
\begin{tabular}{@{}lccccc@{}}
\toprule
\textit{Go2} & Straight & \shortstack{L. Curve} & \shortstack{R. Curve} & \shortstack{Sharp Turn} & \shortstack{Sudden Stop} \\ \midrule
BC & 0.80 & 0.60 & 1.00 & 0.00 & 0.80 \\
Ours & 1.00 & 1.00 & 1.00 & 1.00 & 1.00 \\ \bottomrule
\end{tabular}
\end{subtable}
\vspace{-2em}
\end{table}

\noindent\textbf{Safe Navigation.}  
We provide quantitative results for the safe navigation task in \cref{tab:avoid}, in which the model needs to learn an end-to-end policy with the raw RGBD images as the input. It is clear that BC struggles with most sub-tasks and only achieves a 25\% success rate on moving straight. Our method outperforms BC by a clear margin and performs well even on the most difficult dynamic obstacle avoidance, even the agent is trained in less than 10 minutes. We observe in the experiments that the BC policy is prone to being stuck in unfamiliar situations, such as facing toward the wall. Furthermore, the BC policy would also make emergency stops at locations where it shouldn't stop. We speculate that with an end-to-end policy, BC is more likely to overfit the raw pixel values that are irrelevant to the task and can easily be changed by the lightning. Our method suffers less from overfitting and achieves better safety and efficiency as the human operator can intervene and correct wrong actions during training if any out-of-distribution state is encountered.

\noindent\textbf{Human Following.}
\cref{tab:coco-chase}  shows results for the human following task on different embodiments. Our method performs better than BC under all human behaviors with two different robots. Notably, our method achieves 100\% success rate on the Unitree Go2 robot, where BC fails completely when the human is doing a sharp turn, an action never demonstrated during training. As the human bounding box is not likely to be lost in the demonstration data, the BC agent cannot recover from the out-of-distribution states of lost tracks during the sharp turn. Our method can still infer human direction from the history information by learning from the intervened human demonstrations of these challenging states.

\begin{figure}[!t]
\centering
\includegraphics[width=0.95\linewidth]{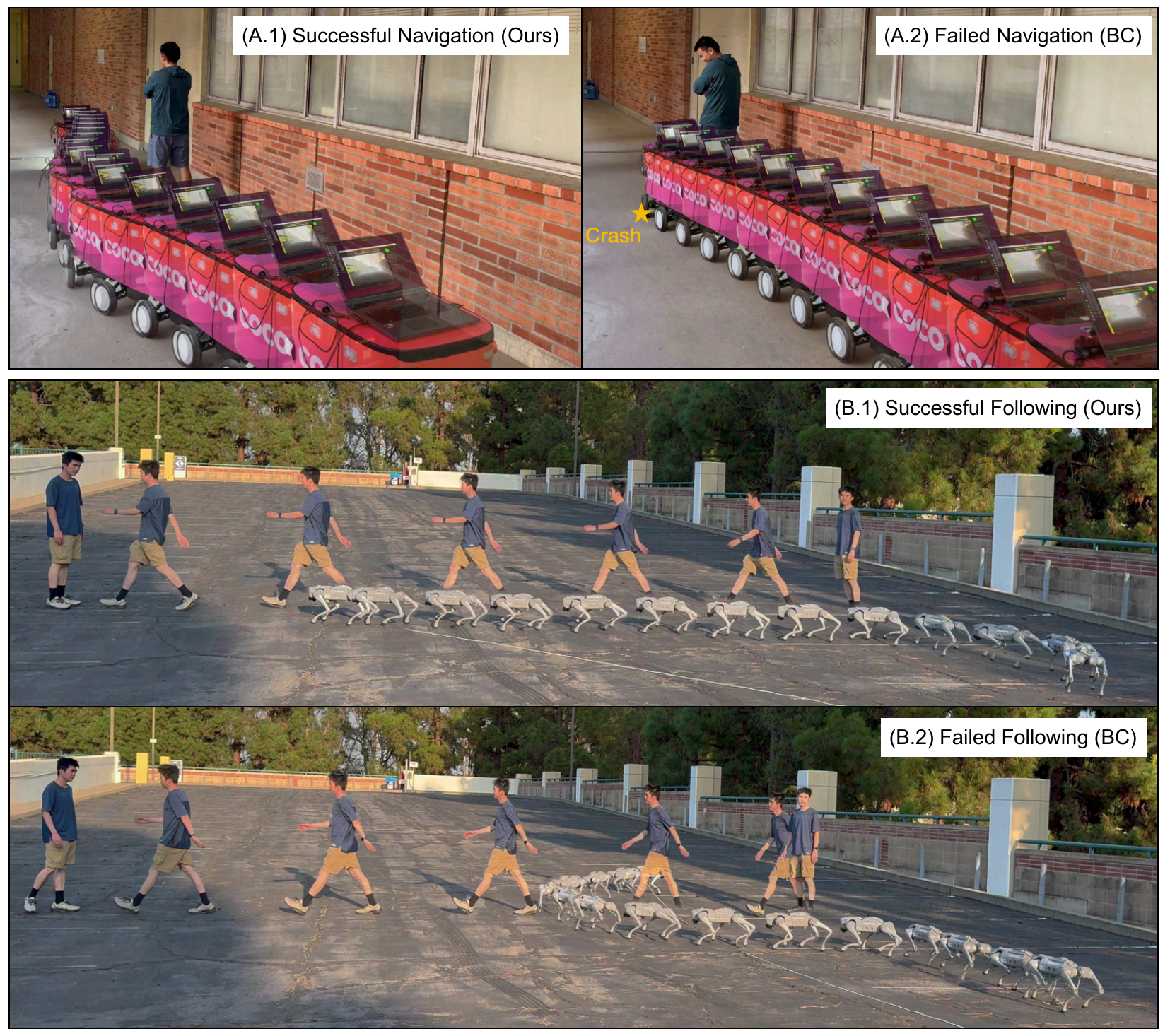}
\vspace{-0.5em}
\caption{
\textbf{Qualitative results for the Safe Navigation and Human Following tasks.}
\textbf{(A)} The stacked trajectories for the ``static obstacle avoidance'' subtask.
\textbf{(B)} The stacked trajectories for the ``sharp turn'' subtask.
}
\vspace{-1.5em}
\label{fig:quantitative}
\end{figure}

\noindent \textbf{Qualitative Results}
As shown in~\cref{fig:quantitative}, in the Safe Navigation task, our method successfully avoids a human standing its path while BC collides with the human. In the Human Following task, BC loses its track and turns in the opposite direction while our method keeps following the person.

\vspace{-0.25em}
\section{Conclusion}
\vspace{-0.25em}

We present a data-efficient human-in-the-loop learning method that combines imitation learning and reinforcement learning to address challenging tasks on mobile robots.
We can train mobile robots to conduct safe navigation and human following within 15 minutes. 
Our method significantly improves performance and safety in challenging scenarios, including sharp turns and sudden stops.
In the simulation experiment, we show that our method can achieve near-optimal performance with fewer human interventions.

\noindent\textbf{Limitation.}
While this work mainly focuses on proving the applicability of the online human-in-the-loop method on real-world robots, a natural extension is to incorporate even more challenging tasks such as long-horizon goal-conditioned navigation~\cite{shah2023gnm}, or manipulation~\cite{luo2023rlif}.

\noindent \textbf{Acknowledgments} This project was supported by NSF grants IIS-2339769 and NSF CCF-2344955. ZP is supported by the Amazon Fellowship via UCLA Science Hub.

\bibliographystyle{IEEEtran}
\bibliography{IEEEfull,root,robot_dog_reference}

\end{document}